\documentclass[balance,colorlinks,upint,subscriptcorrection,varvw,mathalfa=cal=boondoxo]{asmeconf}
\usepackage[utf8]{inputenc}

\hypersetup{%
	pdfauthor={John H. Lienhard},									  
	pdftitle={ASME Conference Paper LaTeX Template},                  
	pdfkeywords={ASME conference paper, LaTeX template, BibTeX style},
	pdfsubject = {Describes the asmeconf LaTeX template},			  
	pdflicenseurl={https://ctan.org/pkg/asmeconf},
}

\usepackage{caption}
\usepackage{array}
\usepackage{tabularx}

\begin{document}


\ConfName{Proceedings of the ASME 2026 21st International \linebreak Manufacturing Science and Engineering Conferences}
\ConfAcronym{MSEC2026}
\ConfDate{June 14-18, 2026} 
\ConfCity{State College, Pennsylvania} 
\PaperNo{MSEC2026-182794}


\title{Evaluating Large and Lightweight Vision Models for Irregular \linebreak Component Segmentation in E-Waste Disassembly} 
 
%
%
%

\SetAuthors{%
    Xinyao Zhang\affil{1}, 
	Chang Liu\affil{2},
    Xiao Liang\affil{2}, 
    Minghui Zheng\affil{2}, 
	Sara Behdad\affil{3}\CorrespondingAuthor{sarabehdad@ufl.edu}
	}

\SetAffiliation{1}{Florida State University, Industrial and Manufacturing Engineering, Tallahassee, FL, USA}
\SetAffiliation{2}{Texas A\&M University, Mechanical Engineering, College Station, TX, USA}
\SetAffiliation{3}{University of Florida, Environmental Engineering Sciences, Gainesville, FL, USA}



\maketitle



\keywords{Electronic Waste, Waste Electrical and Electronic Equipment (WEEE), Semantic Segmentation, Transformer-based Models, Benchmark, Computer Vision for Disassembly}


\begin{abstract}
\vspace{-3pt}

Precise segmentation of irregular and densely arranged components is essential for robotic disassembly and material recovery in electronic waste (e-waste) recycling. This study evaluates the impact of model architecture and scale on segmentation performance by comparing SAM2, a transformer-based vision model, with the lightweight YOLOv8 network. Both models were trained and tested on a newly collected dataset of 1,456 annotated RGB images of laptop components including logic boards, heat sinks, and fans, captured under varying illumination and orientation conditions. Data augmentation techniques, such as random rotation (±15°), flipping, and cropping, were applied to improve model robustness. YOLOv8 achieved higher segmentation accuracy (mAP50 = 98.8\%, mAP50–95 = 85\%) and stronger boundary precision than SAM2 (mAP50 = 8.4\%). SAM2 demonstrated flexibility in representing diverse object structures but often produced overlapping masks and inconsistent contours. These findings show that large pre-trained models require task-specific optimization for industrial applications. The resulting dataset and benchmarking framework provide a foundation for developing scalable vision algorithms for robotic e-waste disassembly and circular manufacturing systems.

\end{abstract}



\section{Introduction}

Accurate robotic disassembly of electronic waste (e-waste) requires precise identification of irregularly-shaped components. Spatial misalignment compromises the ability of robots to determine the position and orientation of parts for practical manipulation. Existing object segmentation approaches struggle to detect non-convex boundaries at the pixel level. 
This study investigates the applicability of Meta’s Segment Anything Model (SAM) for fine boundary detection using an e-waste dataset. To facilitate robust model evaluation and open-source data, the collected dataset combines complex polygonal attachments, such as circuit boards and fans, into a single disassembly layer. A total of 1,456 RGB images are processed by the SAM after applying data augmentation strategies, and the boundary precision is compared against the state-of-the-art YOLOv8 (You Only Look Once) model. The YOLOv8 architecture demonstrates superior accuracy in identifying complex object contours, minimizing boundary errors, and improving part localization. The evaluation shows unique advantages of each architecture in handling object boundaries and scale variations. The study establishes a benchmark dataset that supports future work on irregular component segmentation and model optimization for robotic e-waste disassembly. 
\setlength{\parskip}{0pt}

E-waste component segmentation presents technical difficulties that set it apart from conventional object recognition tasks. Components such as heat sinks and circuit boards exhibit non-convex geometries with irregular boundaries, thin protrusions, and internal cutouts. These shapes differ from the compact objects commonly found in benchmarks such as COCO and Pascal VOC \cite{lin2014microsoft,everingham2010pascal}. Laptop assemblies also contain densely packed components with narrow inter-part gaps, which increases the risk of boundary overlap and occlusion. In addition, components of the same category vary in shape, size, and layout among different manufacturers with high intra-class variation. Visual complexity further intensifies the challenge. Metallic and PCB surfaces produce low-contrast visual appearances with reflective artifacts that reduce the performance of color-based segmentation cues. Practical constraints during disassembly impose additional limitations. During disassembly, components are viewed from limited and fixed angles, which restricts the diversity of training image perspectives. Although similar challenges are shared with related industrial tasks such as PCB inspection \cite{chen2022pcb} and bin picking in manufacturing \cite{mallick2018deep}, the combination of irregular geometry, dense spatial arrangement, and cross-brand variability makes e-waste segmentation a unique benchmark problem.

\section{Literature Review}
\subsection{Irregular Object Segmentation}
As e-waste volumes rise globally, the precise visual identification of electronic components has become a significant bottleneck in automated disassembly and recycling systems. For example, laptop motherboards consist of numerous irregularly shaped components with narrow gaps and complicate robotic segmentation tasks. These components often have fixed orientations, which restrict image capture angles and result in occlusions. Design variations between manufacturers also lead to inconsistencies in component shapes, sizes, and locations. A practical automated solution should detect components of varying sizes and shapes, perform pixel-level segmentation, and delineate clear boundaries even in densely packed arrangements. Object detection uses bounding boxes to identify objects. However, segmentation assigns labels to every pixel in an image and infers object structures in precise spatial awareness for robots \cite{minaee2021image}. Few studies have addressed the segmentation of irregularly shaped e-waste components. Existing methods often lack validation on non-convex objects and fail to achieve robust pixel-level segmentation.

Recent advancements in deep learning improve segmentation and facilitate detailed feature extraction from raw image data. Convolutional neural networks (CNNs) drive progress in related computer vision tasks, and the Mask Region-Convolutional Neural Networks (Mask R-CNNs) have become a helpful tool for instance segmentation \cite{he2017mask}. Chen et al. optimized CNNs to capture dependency features between images and evaluated their applicability to circuit boards analysis \cite{chen2022pcb}. Aarthi et al. applied Mask R-CNNs to detect the waste objects and guide the robot for grasping \cite{aarthi2023vision}. Mask R-CNN structure was also evaluated to detect metal objects and determine their orientations for dual-arm robot grasping \cite{kijdech2024manipulation}. Jahanian et al. \cite{Jahanian_2019_CVPR_Workshops} proposed a modified Mask R-CNN model for multi-task instance segmentation purposes, which detected the boundaries of electric components on PCB boards. Despite these advancements, CNN-based approaches struggle in unstructured environments when dealing with densely packed components where fine details and overlapping structures require feature extraction techniques.

The YOLO framework has gained attention for fine-grained object detection tasks. The YOLO framework is a single-stage CNN architecture that reformulates object detection as a regression problem. Two-stage CNN methods such as Mask R-CNN first generate region proposals and then classify them. YOLO instead predicts bounding boxes and class probabilities simultaneously in a single forward pass. This design reduces inference time and makes YOLO suitable for real-time applications. Developments in YOLOv5 and YOLOv8 improved small-object detection in multi-camera vision systems \cite{zhou2024towards}. For example, a YOLOv5 model was designed for image segmentation in scenarios involving single circular boards \cite{gluvcina2023detection}. Also, Zhang et al. demonstrated its industrial applicability by optimizing YOLOv4 to detect screws in various laptop brands \cite{zhang2023automatic}.  However, since YOLO focuses on object detection, hybrid approaches that combine detection with fine-grained segmentation are necessary to achieve precise boundary delineation \cite{mallick2018deep}.

More recently, Transformer-based architectures emerge as promising solutions for segmentation tasks. The SEgmentation TRansformer (SETR) replaced traditional CNNs with a Transformer encoder to achieve pixel-wise segmentation \cite{zheng2021rethinking}. SegLater introduced a lightweight design, making Transformer-based segmentation more practical for real-world applications \cite{xie2021segformer}. In waste recycling, Dong et al. applied a Transformer model to classify construction waste components in complex onsite environments \cite{dong2022computer}. Despite strong global feature representation performance, the high computational cost of Transformers limits their real-time applicability in robotic systems. 

Despite significant progress in segmentation research, robust pixel-level methods for irregular and densely arranged components are limited. This study aims to address this gap by evaluating and comparing advanced vision architectures for accurate component segmentation in robotic e-waste disassembly.

\subsection{E-waste Dataset}
Given the importance of datasets in the development of smart e-waste recycling systems, previous studies have collected datasets suitable for performing object detection and computer vision tasks in this domain. However, most of them focus on detecting e-waste in solid waste streams.

IIiev et al. introduced an e-waste dataset consisting of 19,613 images for 77 different electronic device classes \cite{iliev2024proposal}. This dataset, publicly available, was assembled by combining 91 open-source datasets from the Roboflow Universe, supplemented by manually annotated images from Wikimedia Commons. Its primary purpose was to support e-waste detection. Similarly, Voskergian and Ishaq collected around 4,000 images of four electronic object types from the Open Images Dataset V7 and evaluated YOLO object detection models on this dataset \cite{voskergian2023smart}. Zhou et al. developed a dataset with 660 images specifically for screw hole detection on laptops, using a combination of YOLO and Faster R-CNN architectures \cite{zhou2024towards}. TrashNet \cite{yang2016classification} contains seven categories, including plastic, glass, metal, etc. It provides 2527 images for training waste sorting and management models \cite{lu2022computer}. Similarly, Ekundayo et al. \cite{ekundayo2022device} also proposed a WasteNet dataset, which is a combination of multiple waste datasets with a total of $33,520$ images.

Madhav et al. proposed a dataset containing 8,000 images for eight consumer electronic device classes, aimed at classification tasks; however, the dataset is not publicly available \cite{shreyas2022application}. Selvakanmani et al. utilized two public datasets from Kaggle namely, the Starter E-Waste Dataset and the Compressed E-Waste Dataset, which contain 613 and 806 images, respectively. They evaluated deep learning models for classification tasks on these datasets \cite{selvakanmani2024optimizing}. Liu et al. proposed an automated disassembly system and a customized phone dataset, which contains over 2,300 images of different phone layer components \cite{liu2026raise}. Baker et al. collected 650 images representing 12 smartphone classes, which were used for waste sorting experiments \cite{abou2021transfer}. Islam et al. presented 1,053 images of eight different electronic devices and implemented Transformer methods for e-waste classification \cite{islam2023ewastenet}. Rupok et al. introduced a dataset containing 1,004 images categorized into four classes, though this dataset is not publicly available \cite{rupok2024electrosortnet}. Zhou et al. evaluated a dataset with 52,800 multi-view images for 22 classes from a recycling facility and applied a graph transformer model for e-waste classification \cite{zhou2025multi}. 

Quang et al. collected disassembled images of 10 different cellphones, containing 11 component classes \cite{jahanian2019see}. In total, they annotated 533 images, which were used for instance segmentation and detection tasks. Also, Wu et al. evaluated a dataset with 513 color and depth images of laptops for defect detection and segmentation \cite{wu2024color}. 

\subsection{SAM Segmentation}
The Segment Anything Model represents a shift toward foundation models for image segmentation \cite{kirillov2023segment}. SAM was trained on the SA-1B dataset, which contains over 11 million images and 1 billion masks. The model uses a promptable design that accepts points, boxes, or text as input and produces class-agnostic segmentation masks. SAM demonstrates strong zero-shot generalization for diverse visual domains, including medical imaging \cite{ma2024segment} and remote sensing \cite{chen2024rsprompter}. However, SAM faces challenges when applied to specialized industrial datasets without domain-specific fine-tuning. Objects with low contrast, overlapping boundaries, or unusual shapes can degrade its segmentation quality.

SAM2 extends the original SAM to support video segmentation \cite{ravi2024sam}. SAM2 introduces a streaming architecture with a memory attention module that tracks objects over frames. The model was trained on the SA-V dataset, which adds video annotations to the image-based SA-1B data. SAM  processes each image independently, however SAM2 maintains a memory bank that stores features from previous frames to improve temporal consistency. When applied to static images, SAM2 retains the encoder-decoder structure of SAM but benefits from a refined training process and improved mask decoder. In this study, we evaluate SAM2 on static e-waste images to assess whether these architectural updates lead to higher segmentation accuracy for irregular components.

\section{Methodology}
\subsection{SAM2 Architecture}
\begin{figure*}[!t]
  \centering
  \includegraphics[width=\linewidth]{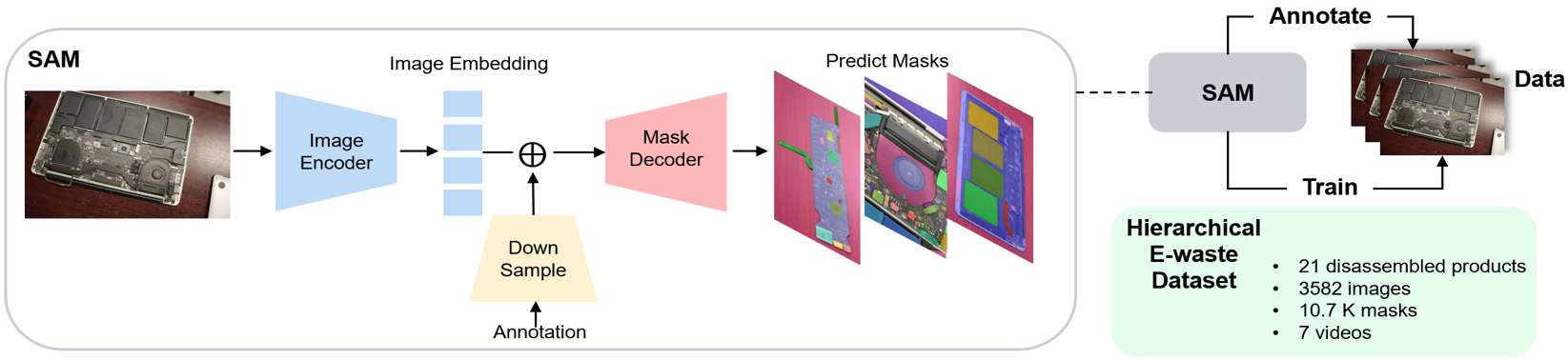}
  \caption[SAM2 architecture]{SAM2 architecture.}
  \label{fig:1}
\end{figure*}

E-waste disassembly consists of handling a diverse array of components, from tiny screws to large circuit boards. These objects often appear in layered configurations, which require fine-grained segmentation of complex shapes and overlapping parts. To facilitate this, we aim to compare the performance of a large vision model with a lightweight state-of-the-art YOLO model. 
The proposed SAM2 architecture, shown in Figure~\ref{fig:1}, builds upon the original SAM to better accommodate the complexities of e-waste disassembly. The original SAM model has demonstrated considerable success in fields ranging from autonomous driving to medical imaging, however, it faces challenges when segmenting irregularly shaped and diminutive components. SAM2 addresses these issues by introducing a hierarchical encoder-decoder framework and a generative learning component, both of which facilitate more precise segmentation of e-waste objects that vary greatly in size, shape, and texture.

This hierarchical design addresses the diversity of e-waste disassembly scenarios. At lower encoder levels, the model extracts fine spatial details such as edges and corners. These features are important for delineating small components, including IO boards and connector clips, which occupy only a small portion of the image. At higher levels, the encoder captures broader contextual information, such as the spatial relationships between adjacent parts. These relational features help distinguish overlapping components in dense assemblies, for example, when a heat sink partially covers a logic board. The decoder combines these multi-scale representations through progressive upsampling. This fusion supports the model to handle the large size variation present in e-waste images, where logic boards may span half the image and small connectors cover fewer than 100 pixels.

The goal of SAM2 is to capture hierarchical structures more precisely than traditional dense classification models. SAM2 extends the large-scale pretraining of SAM by using additional hierarchical annotations and a generative decoding strategy, where it can segment irregular objects while preserving high-level contextual understanding.

Let $x \in \mathbb{R}^{H \times W \times 3}$ denote an input image. SAM2 encodes $x$ at multiple scales to capture both fine-grained details and coarse contextual cues. Formally, we define:
\begin{equation}
E(x) = \{E_1(x), E_2(x), \dots, E_n(x)\},
\end{equation}
where $E_i(x)$ represents the features extracted at the $i$-th level of the encoder. Lower levels ($i$ small) capture high-resolution details, while higher levels ($i$ large) encode more global information. This hierarchical approach is useful for e-waste scenarios, where small objects (e.g., screws) and larger assemblies (e.g., motherboards) coexist in the same scene.

The decoder reconstructs the image representation and produces segmentation masks by combining features from multiple scales:
\begin{equation}
D(E(x)) = \sum_{i=1}^{n} D_i\bigl(E_i(x)\bigr).
\end{equation}
Each decoder block $D_i(\cdot)$ performs upsampling and feature reconstruction from its paired encoder layer. The design retains critical local structures, including edges and corners, and incorporates contextual cues from higher-level representations.

The mask prediction module applies a nonlinear transformation to the decoder output to generate a semantic segmentation mask:
\begin{equation}
M(x) = \sigma\bigl(D(E(x))\bigr),
\end{equation}
where $\sigma(\cdot)$ denotes the sigmoid activation function. This step produces a pixel-wise probability map that indicates the likelihood of each pixel belonging to a specific object class. By learning to segment a wide variety of e-waste components, SAM2 can handle complex boundaries and overlapping parts that often arise in disassembly tasks.

SAM2 incorporates a generative learning paradigm to modify its segmentation outputs further. In scenarios with incomplete or ambiguous annotations, the model uses learned latent representations to infer plausible object shapes and boundaries. This capability is particularly advantageous for e-waste disassembly, where items may be partially occluded or exhibit non-uniform geometries. SAM2 synthesizes high-quality segmentations from dense embeddings, and extends large-scale pretraining toward domain-specific e-waste segmentation tasks.

Through its hierarchical encoder-decoder design and generative learning mechanism, SAM2 adjusts to diverse e-waste scenarios. It maintains robust performance on both large, complex assemblies and small, irregular components. Empirical results indicate that SAM2’s approach to multi-scale feature extraction and generative inference improves segmentation quality, which further makes it a promising solution for automated disassembly systems in industrial settings.

\subsection{YOLOv8 Architecture}
\begin{figure*}[!t]
  \centering
  \includegraphics[width=0.8\linewidth]{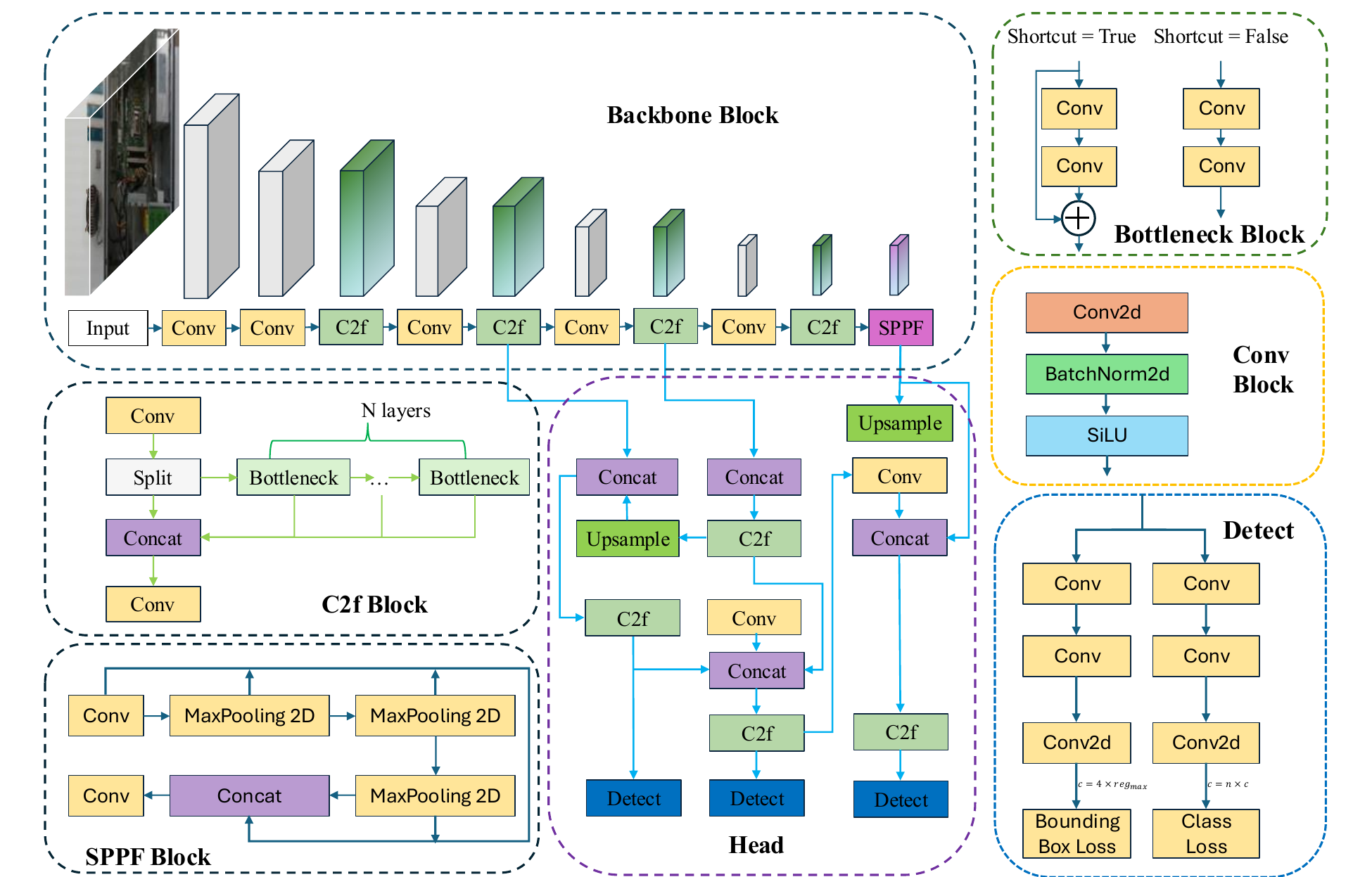}
  \caption[Yolov8 architecture]{Yolov8 architecture}
  \label{fig:yolo}
\end{figure*}

YOLOv8 provides a unified architecture for segmentation that incorporates feature fusion, anchor-free detection, and multi-scale feature learning. The structure of YOLOv8 can be found in Figure~\ref{fig:yolo}. The framework consists of three principal components: a CSPDarknet53 backbone augmented with a C2f module, an anchor-free decoupled segmentation head, and a hybrid loss function design.

The backbone extracts essential features using a 53-layer deep convolutional neural network with a cross-stage partial network (CSP) architecture. The CSP uses residual connections to the backbone and direct connections to the segmentation head part. Another significant improvement in YOLOv8 is the use of a C2f module (cross-stage partial bottleneck with two convolutions) instead of a traditional C3 convolutional module, which enhances accuracy and reduces inference time. The C2f module employs split-and-merge operations to aggregate multi-scale features and improve the capture of fine-grained spatial patterns for irregular shapes. YOLOv8 also uses a smaller 3x3 convolutional layer to replace the original 6x6 convolutional layer in YOLOv5 in the backbone network. The module facilitates the recognition of fragmented objects, which are commonly encountered in e-waste collection, by integrating shallow-layer features with deeper representations. 

An anchor-free strategy, employed by YOLOv8, facilitates the direct regression of bounding box centers and sizes. Instead of predefined anchor boxes, the model directly predicts bounding box coordinates and adjusts to the heterogeneous shapes and scales. A decoupled task-specific head segregates detection confidence, classification, and localization into parallel processing streams. This separation minimizes cross-task interference in cluttered scenes where overlapping or fragmented objects are prevalent.

For segmentation, a dual-resolution fusion strategy is employed: high-resolution features from the backbone preserve fine structural details, while upsampled semantic features from the C2f module provide contextual awareness. These complementary data streams are combined within a lightweight mask head, which synthesizes pixel-accurate boundaries through hierarchical feature aggregation. The result is robust contour delineation, even for objects with discontinuous edges or reflective surfaces in the e-waste dataset. 

Segmentation performance is optimized using a hybrid loss function defined as: 
\begin{equation} L_{\text{total}} = \lambda_{\text{CIoU}} L_{\text{CIoU}} + \lambda_{\text{DFL}} L_{\text{DFL}} + \lambda_{\text{BCE}} L_{\text{BCE}} \end{equation} 
where $L_{\text{CIoU}}$ (complete intersection over union loss) controls bounding box localization by accounting for overlap, aspect ratio, and center distance between predicted and ground-truth regions. $L_{\text{DFL}}$ (distribution focal loss) sharpens probabilistic predictions by penalizing uncertain classifications in ambiguous regions. $L_{\text{BCE}}$ (binary cross-entropy loss) refines pixel-wise mask boundaries by minimizing errors between predicted and true segmentation maps. The weighting coefficients ($\lambda_{\text{CIoU}}$, $\lambda_{\text{DFL}}$, $\lambda_{\text{BCE}}$) determine the contributions of each term during training. This hybrid formulation addresses challenges in e-waste segmentation, such as irregular shapes, occlusions, and fragmented edges.

\subsection{Comparison of Model Approaches}
SAM2 and YOLOv8 differ in their design objectives and segmentation strategies. SAM2 is a foundation model trained on large-scale and diverse data to produce class-agnostic masks. It generates multiple candidate segmentation masks per image and uses external prompts or post-processing to associate masks with specific classes. This design provides flexibility in representing diverse object structures but may produce redundant or overlapping masks when applied to domain-specific tasks without further optimization.

YOLOv8 is a task-specific architecture that performs detection and segmentation in a single forward pass. It is trained end-to-end on a target dataset and predicts class-labeled masks. The anchor-free detection head and feature fusion module help the model learn domain-specific shape patterns, which benefits the segmentation of irregular components in e-waste images. Table \ref{cp7tab:2} summarizes the main differences between the two models.

\begin{table}[h]
\renewcommand{\arraystretch}{1.2}
\caption{Comparison of SAM2 and YOLOv8 for e-waste segmentation.}
\label{cp7tab:2}
\begin{tabularx}{\linewidth}{@{}p{1.6cm}>{\raggedright\arraybackslash}X>{\raggedright\arraybackslash}X@{}}
\hline
\textbf{Aspect} & \textbf{SAM2} & \textbf{YOLOv8} \\
\hline
Design objective & General-purpose and class-agnostic segmentation & Task-specific detection and segmentation \\
Training strategy & Pre-trained on SA-V dataset, and fine-tuned on target data & Trained from scratch on target dataset \\
Mask generation & Multiple candidate masks per image & One mask per detected object \\
Class assignment & Requires post-processing or prompting & Directly predicts class labels \\
Irregular shapes & Uses learned shape priors from diverse pre-training & Leans domain-specific features \\
Inference speed & Higher computational cost & Lightweight and real-time capable \\
\hline
\end{tabularx}
\end{table}

\section{Dataset}
Data were collected during the manual disassembly of end-of-use laptops. After removing the back case, operators disassembled components layer-by-layer. Images were captured from multiple angles before and after disassembly, each depicting irregularly shaped e-waste components. Depending on the laptop’s condition, images contain either single or multiple components within a layer.
\begin{figure}[!hbt]
  \centering
  \includegraphics[width=\linewidth]{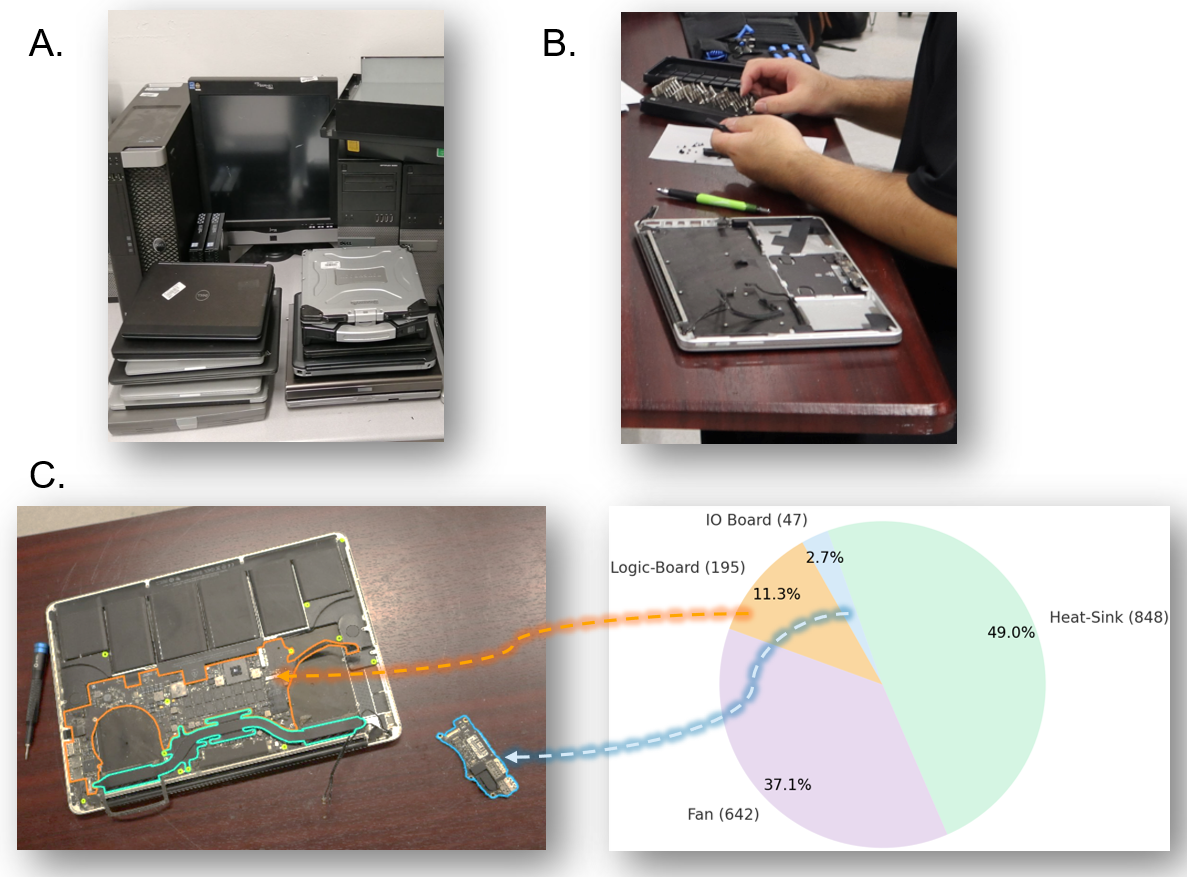}
  \caption[Manual disassembly of e-waste dataset]{Manual disassembly of e-waste dataset.}
  \label{fig:f2}
\end{figure}

The dataset comprises 560 raw images from 3 Apple MacBook models, cleaned and annotated to include 4 e-waste categories: logic boards, heat sinks, IO boards, and fans. Component counts and distributions are shown in Figure~\ref{fig:f2}. The dataset reflects real-world complexity, with objects ranging from large logic boards to small IO boards. A total of 1,717 segmentation masks were annotated by two experts for precise boundary delineation.

For model training, we apply a set of data augmentation strategies that include horizontal and vertical flips, random cropping (0\% minimum zoom and 14\% maximum zoom), and rotations ranging from -15° to +15° without reducing image resolution. Each training image generated three variants, expanding the training set from 448 to 1,344 images. This augmentation enriches the variety of object appearances and strengthens a model’s learning ability. The validation and testing sets each contain 56 images. The dataset provides a benchmark for evaluating segmentation algorithms. 

\section{Results}
\subsection{SAM Segmentation Results} 

The results illustrate SAM performance on the discussed dataset. Experiments compare models without fine-tuning and those fine-tuned at 40, 100, and 200 epochs. Trials also assess low-resolution versus high-resolution images. The evaluations show the influence of fine-tuning and image quality on segmentation accuracy for complex e-waste components.

For fine-tuning, we froze the image encoder of SAM2 and updated only the mask decoder. This strategy reduces computational cost and concentrates gradient updates on domain-specific segmentation behavior. We used the AdamW optimizer with a weight decay of 0.01 and a batch size of 4. We saved model checkpoints at 40, 100, and 200 epochs to examine how segmentation quality evolves with training duration. The loss function combined binary cross-entropy and dice loss to balance pixel-level accuracy and region-level overlap. We also compared fine-tuning with low-resolution and high-resolution images to evaluate the effect of input quality on segmentation performance.

The model processes the entire image and assigns a segmentation label to every pixel. The pretrained SAM model uses a Transformer-based encoder-decoder architecture. This design generates dense segmentation masks in the image. The architecture, trained on large-scale datasets with over 80 million parameters, learns a broad set of visual features and object boundaries. The segmentation map covers every pixel by associating learned representations with spatial regions. The model generalizes from extensive pretraining, where it encountered diverse object shapes, textures, and contexts. Even without task-specific adaptation, the SAM model produces segmentation outputs for the entire input image.

\begin{figure}[!hbt]
  \centering
  \includegraphics[width=\linewidth]{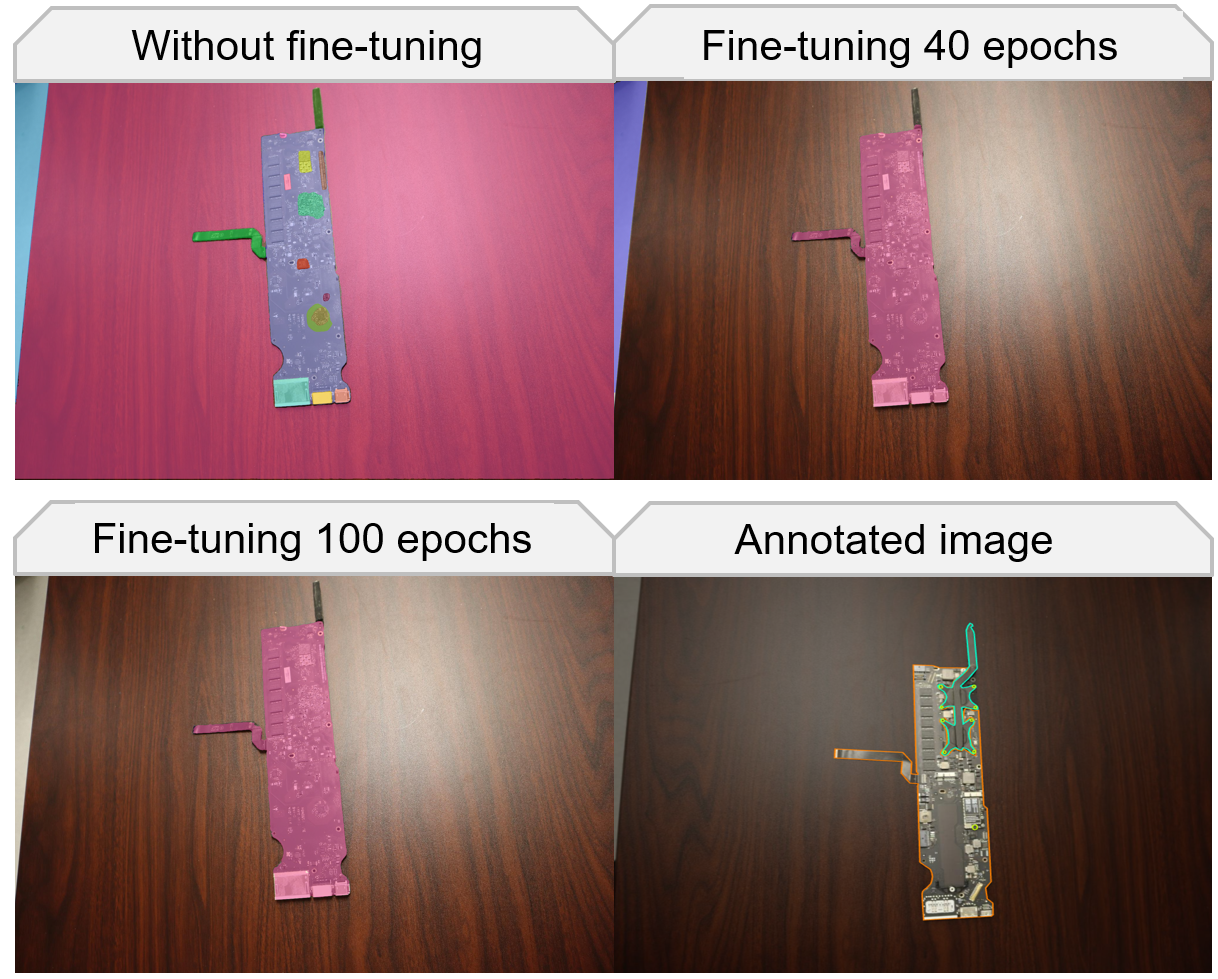}
  \caption[SAM2 instance segmentation for logic board]{SAM2 instance segmentation for logic board.}
  \label{fig:3}
\end{figure}

In Figure~\ref{fig:3}, the model shows improvement after 40 epochs of fine-tuning. The model removes large background regions. At 100 epochs, the model delineates logic board outlines with precision and separates the foreground from the background. The progression demonstrates the benefit of extended fine-tuning in distinguishing target components from background features.

\begin{figure}[!hbt]
  \centering
  \includegraphics[width=\linewidth]{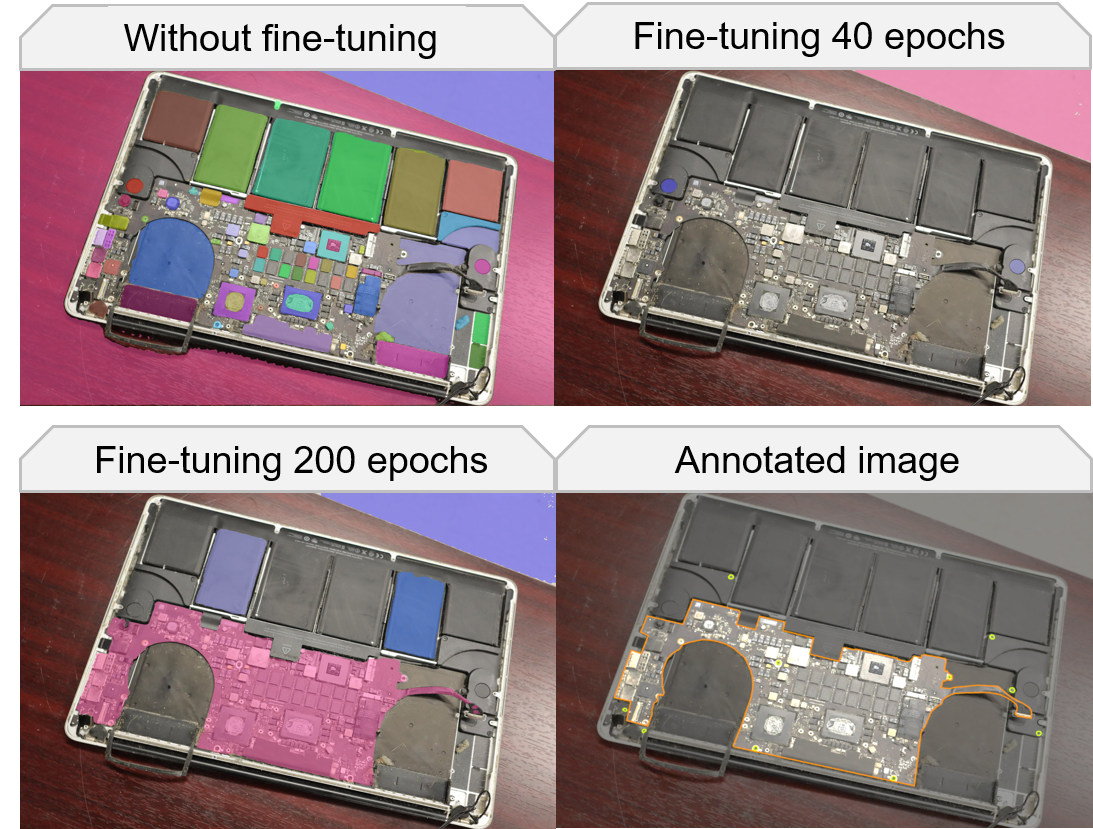}
  \caption[SAM2 instance segmentation for logic board attached at a laptop]{SAM2 instance segmentation for logic board attached at a laptop.}
  \label{fig:4}
\end{figure}

\begin{figure}[!hbt]
  \centering
  \includegraphics[width=\linewidth]{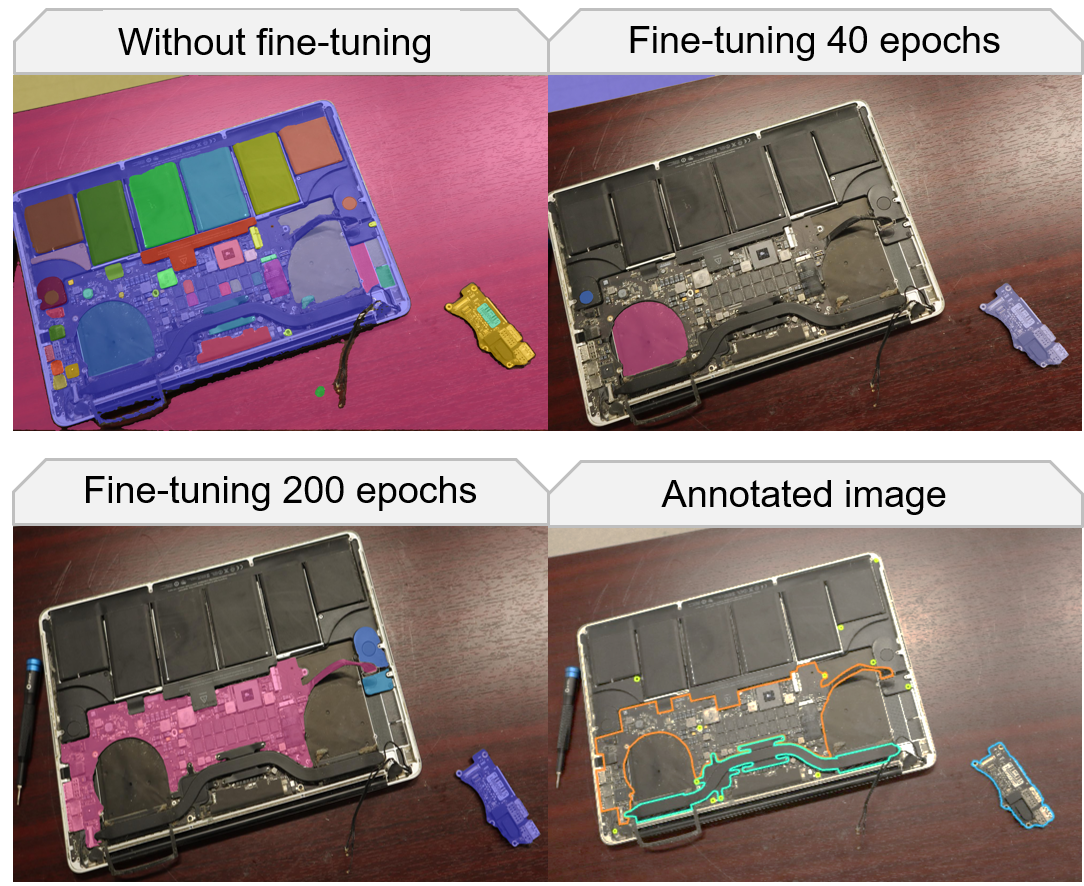}
  \caption[SAM2 instance segmentation for logic board, heat sink and IO board]{SAM2 instance segmentation for logic board, heat sink and IO board.}
  \label{fig:5}
\end{figure}

Figure~\ref{fig:4} and \ref{fig:5} present performance for single-object and multi-object segmentation, respectively. In Figure~\ref{fig:4}, the model without fine-tuning assigns random segmentation labels. At 40 epochs, the attached logic board remains undetected. At 200 epochs, the segmentation captures the primary object with defined boundaries. This result suggests that 200 epochs may suffice for single-object segmentation tasks. 

On the other hand, Figure~\ref{fig:5} shows that at 200 epochs, the model detects two objects but fails to detect the heat sink. For multi-object segmentation, training for 200 epochs does not capture all pre-labeled components. The results suggest that further fine-tuning or architecture adjustment may be required for scenarios with dense objects or varied visual characteristics.

\begin{figure}[!hbt]
  \centering
  \includegraphics[width=\linewidth]{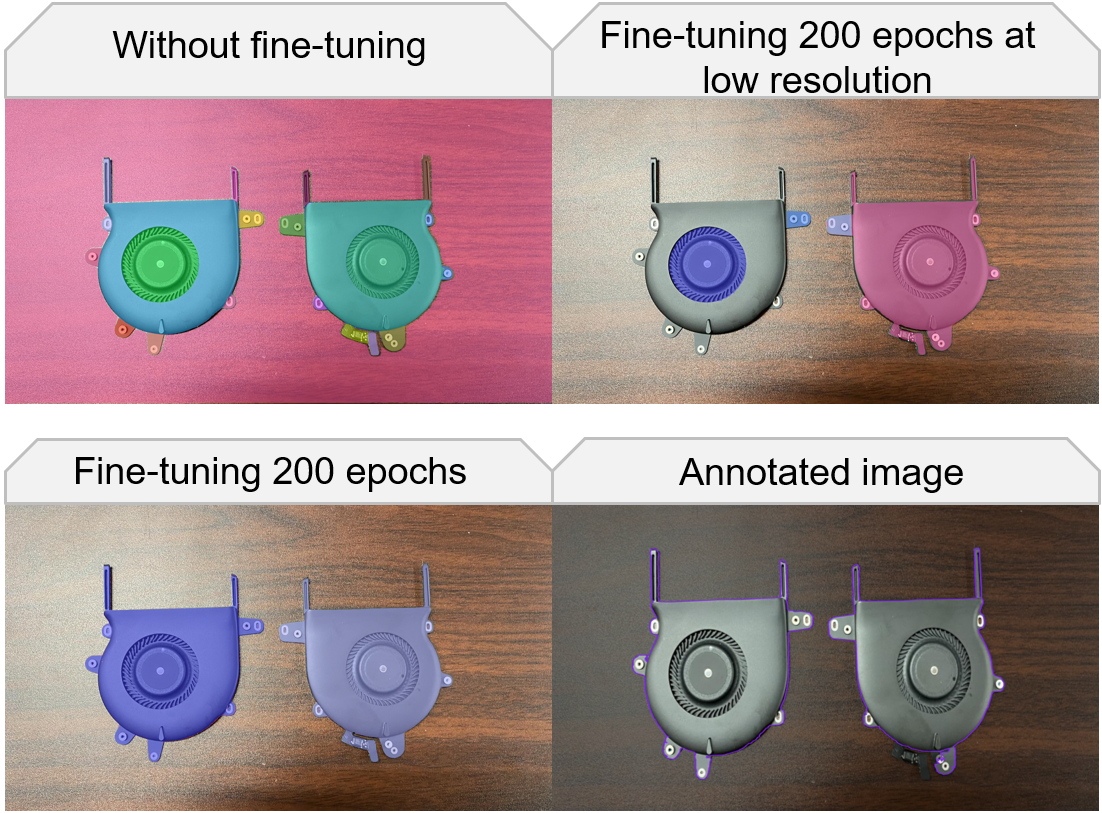}
  \caption[SAM2 instance segmentation for double-fan]{SAM2 instance segmentation for double-fan. All images use high-resolution input by default unless otherwise noted.}
  \label{fig:6}
\end{figure}

Figure~\ref{fig:6} illustrates the influence of image resolution on segmentation performance. Without fine-tuning, the model produces segmentation masks for both targets and background. Fine-tuning with low-resolution images improves segmentation but task-specific capture. Fine-tuning with high-resolution images yields coherent boundaries and improved object separation. The observation underscores the importance of high-quality data for e-waste disassembly tasks that include small and intricate components.

\subsection{YOLOv8 Segmentation Results}
\begin{figure}[!hbt]
  \centering
  \includegraphics[width=\linewidth]{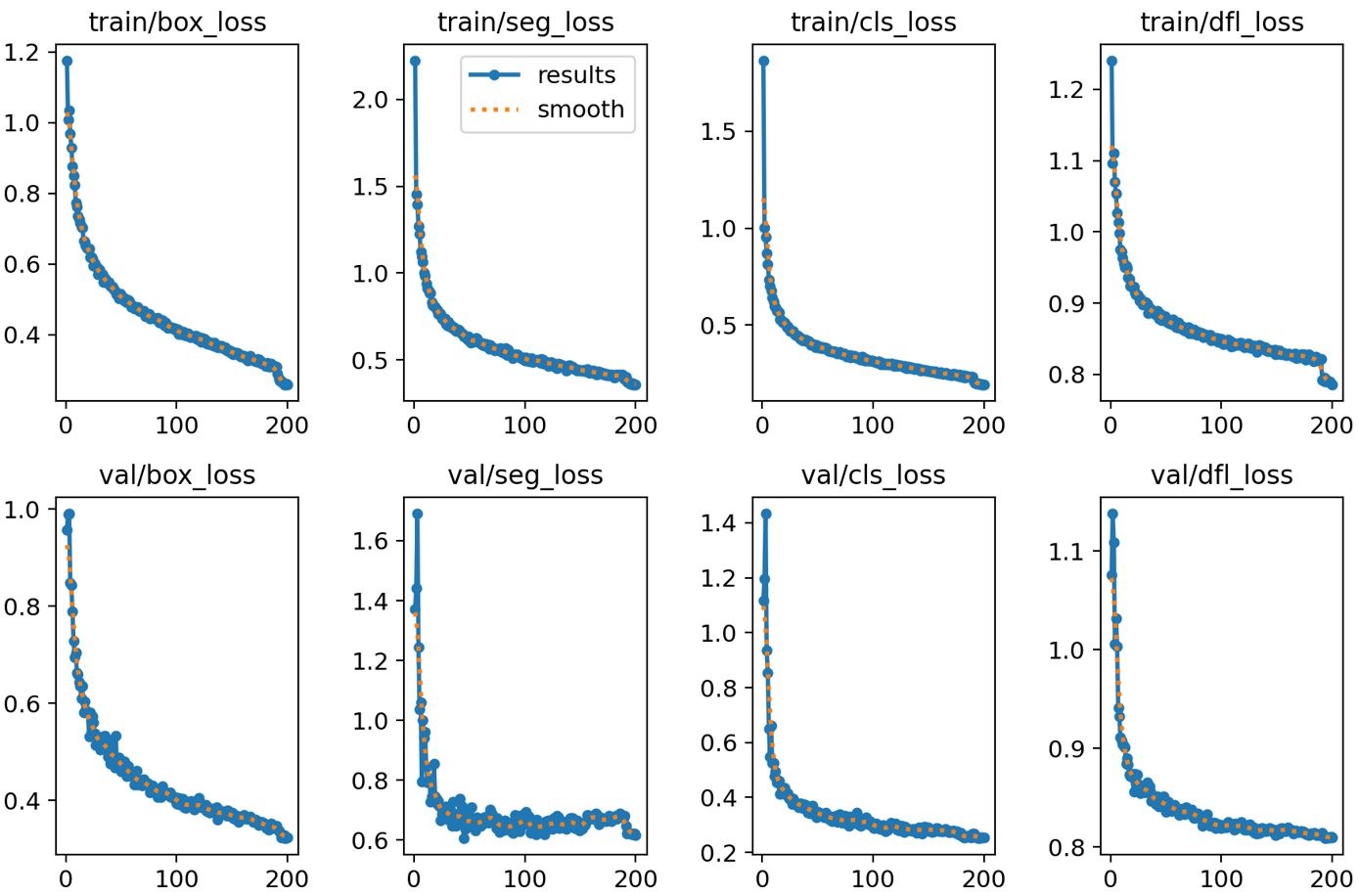}
  \caption[YOLOv8 training and validation loss]{YOLOv8 training and validation loss.}
  \label{fig:7}
\end{figure}

The results plot (in Figure~\ref{fig:7}) shows training and validation losses for box regression, segmentation accuracy, classification accuracy, and distribution focal loss. With increasing epochs, both training and validation losses decrease steadily, indicating reduced overfitting.

Box loss governs the accuracy of predicted bounding box coordinates. Its decreasing trend reflects improved spatial localization of e-waste components. Segmentation loss measures mask boundaries and pixel-level accuracy. Lower values correspond to more precise delineation of irregular shapes. Classification loss distinguishes foreground from background and reduces false positives. Distribution focal loss sharpens prediction probabilities in ambiguous regions and results in reliable object center predictions.

The four loss curves start at different values and converge to different endpoints. In particular, segmentation loss is slightly above the other losses and reflects the challenge of capturing fine-grained boundaries for irregular shapes.

\begin{figure}[!hbt]
  \centering
  \includegraphics[width=\linewidth]{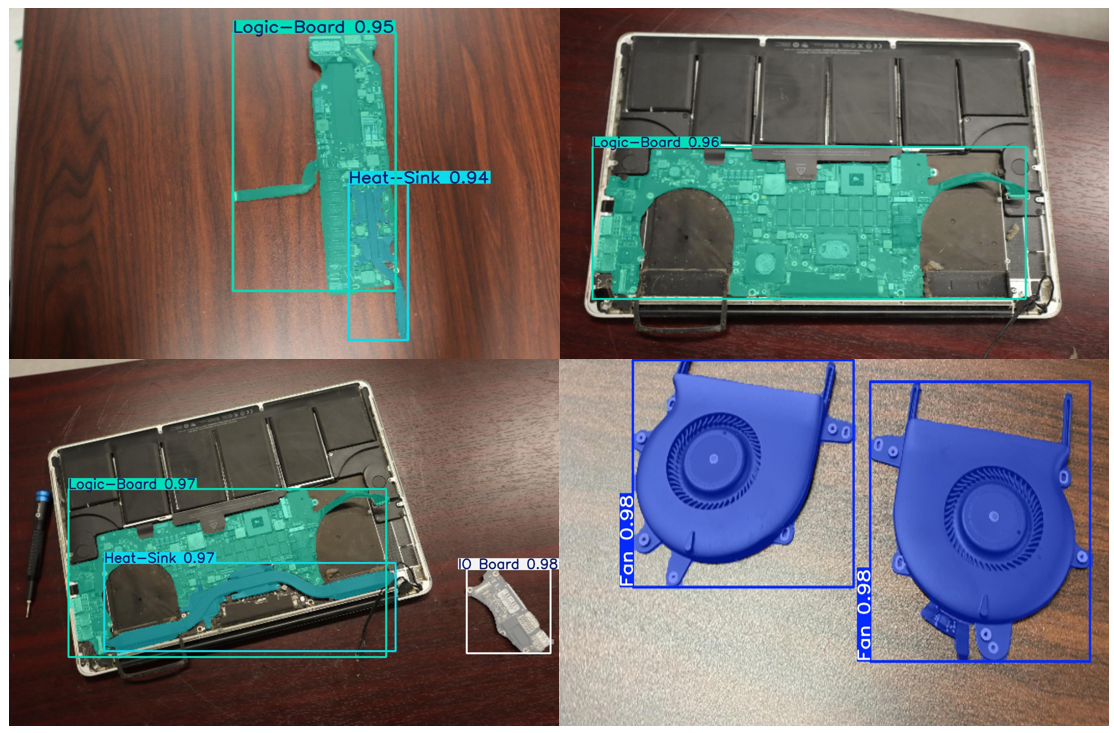}
  \caption[YOLOv8 instance segmentation for four components]{YOLOv8 instance segmentation for four components.}
  \label{fig:8}
\end{figure}

The YOLOv8 segmentation output in Figure~\ref{fig:8} includes instance segmentation with bounding boxes and predicted classes for four e-waste components. The confidence score reflects the model's certainty that each detected object belongs to its assigned class. In YOLOv8, confidence scores derive from two components: the objectness score and the class probability. The objectness score represents the probability that a bounding box contains any object, while the class probability indicates the likelihood that the detected object belongs to a specific class. The final confidence score is calculated as the product of these two values:
\begin{equation}{Confidence} = \text{objectness score} \times \text{class probability} \end{equation}

\subsection{Segmentation Model Performance Comparison}
To compare the models, this study adopts evaluation metrics from the standard COCO segmentation framework \cite{lin2014microsoft}. mAP\textsubscript{50} calculates average precision at an intersection over union (IoU) threshold of 0.5, while mAP\textsubscript{50-95} combines average precision over thresholds from 0.5 to 0.95. mAP\textsubscript{50} is adequate for tasks that allow minor inaccuracies, whereas mAP\textsubscript{50--95} is critical for scenarios requiring precise object delineation.

\begin{table}[h]
\renewcommand{\arraystretch}{1.2} 
\newcolumntype{L}{>{\raggedright\arraybackslash}X}
\newcolumntype{C}{>{\centering\arraybackslash}X}
\caption{Segmentation results of two models \label{cp7tab:1}}
\begin{tabularx}{\linewidth}{@{\extracolsep{\fill}}lll@{}}
\hline
Model & mAP\textsubscript{50} & mAP\textsubscript{50-95} \\
\hline
SAM2 & 8.4\% & 9\% \\
YOLOv8 & 98.8\% & 85\% \\
\hline
\end{tabularx}
\end{table}

Table~\ref{cp7tab:1} presents segmentation performance for SAM2 and YOLOv8. YOLOv8 achieves mAP$_{50}$ of 98.8\% and mAP$_{50\text{--}95}$ of 85\%. SAM2 achieves mAP$_{50}$ of 8.4\% and mAP$_{50\text{--}95}$ of 9\%. The large gap in quantitative scores does not indicate that SAM2 fails to segment e-waste components. As shown in Figure~\ref{fig:3} to \ref{fig:6}, SAM2 produces meaningful segmentation after fine-tuning, including defined logic board contours in Figure~\ref{fig:3}, primary object capture in single-object scenes in Figure~\ref{fig:4}, and improved object separation with high-resolution input in Figure~\ref{fig:6}. The low mAP scores instead reflect a mismatch between SAM2's class-agnostic mask generation strategy and the COCO evaluation protocol, which requires correct class-mask association. SAM2 generates multiple candidate masks per image without assigning class labels. Many of these masks are counted as false positives even when they correspond to visually meaningful regions. SAM2 also produces overlapping masks for adjacent components in dense scenes of Figure~\ref{fig:5}, which further increases false positive counts. On the other hand, YOLOv8 directly predicts class-specific masks with non-maximum suppression, which eliminates redundant detections. The similar mAP$_{50}$ and mAP$_{50\text{-}95}$ values for SAM2 indicate that its performance is consistently low for all IoU thresholds. This confirms that SAM2's primary limitation is not boundary precision but rather the lack of class-specific mask assignment. If boundary quality were the main issue, a larger drop from mAP$_{50}$ to mAP$_{50\text{-}95}$ would be expected, as observed with YOLOv8 (98.8\% to 85\%).

The choice of mAP\textsubscript{50} and mAP\textsubscript{50-95} follows the standard COCO segmentation framework \cite{lin2014microsoft}, which supports comparison with published results. However, these metrics are designed for class-aware evaluation and may not fully reflect the quality of a class-agnostic model such as SAM2. We also note that SAM2 and YOLOv8 serve different purposes: SAM2 is a general-purpose foundation model, YOLOv8 is task-specific. The purpose of this comparison is to assess their practical suitability for e-waste segmentation. The results suggest that task-specific models achieve higher scores under standard protocols. Foundation models require additional post-processing or prompt engineering to translate their segmentation capability into class-aware predictions.

The test set contains 56 images from three Apple MacBook models. The limited device diversity and consistent component layout within MacBook models may contribute to YOLOv8's high mAP$_{50}$ score. We consider the current results a strong baseline for the proposed dataset. Future work will extend the evaluation to additional laptop brands and device categories to assess cross-domain generalization.

\section{Conclusion}
In this study, we evaluated two state-of-the-art segmentation algorithms, SAM2 and YOLOv8, using a newly developed dataset of annotated laptop components from e-waste disassembly. The dataset captures real-world complexity through irregular geometries, overlapping parts, and dense spatial arrangements. It could serve as a benchmark for assessing segmentation performance in industrial recycling environments.

YOLOv8 achieved higher segmentation accuracy (mAP\textsubscript{50} = 98.8\%, mAP\textsubscript{50-95} = 85\%) and delivered stable, real-time performance due to its lightweight architecture and feature fusion mechanism. On the other hand, SAM2, a transformer-based model with generative and hierarchical encoding, exhibited flexibility in handling diverse object structures but struggled with precise boundary delineation and produced redundant masks in densely packed scenes. These results show that large, pre-trained models require targeted fine-tuning and task-specific optimization to achieve pixel-level precision in robotic disassembly tasks.

This study establishes a dataset and benchmarking framework to guide future segmentation research in circular manufacturing systems. Limitations include the dataset’s focus on laptop components and restricted fine-tuning iterations for SAM2 due to computational constraints. Future work will extend the dataset to additional device categories, incorporate multimodal sensing (depth and infrared), and investigate hybrid transformer-convolutional architectures optimized for precise, real-time robotic disassembly.


\section*{Acknowledgments}
This work has been supported by the National Science Foundation - USA  under grants CMMI-2349178, CMMI-2026276 and CMMI-2422826. Any opinions, findings, and conclusions or recommendations expressed in this material are those of the authors and do not necessarily reflect the views of the National Science Foundation.



\bibliographystyle{asmeconf}  
\bibliography{asmeconf-sample}


\end{document}